%% file: main.tex
\documentclass[letterpaper]{article}
\usepackage{aaai}
\usepackage{times}
\usepackage{helvet}
\usepackage{courier}
\makeatletter
\def\copyright@year{}
\def\copyright@text{}
\def\copyright@on{}

\def\copyrighttext#1{}
\def\copyrightyear#1{}
\makeatother

\usepackage{booktabs} 
\usepackage{indentfirst}
\usepackage{amsmath}
\usepackage{amssymb}
\usepackage[linesnumbered,algoruled,boxed,lined]{algorithm2e}
\usepackage{tabularx}
\usepackage{array}
\newcommand{\mlf}{\texttt{MLFriend}\xspace}
\usepackage{subcaption}
\newcommand{\cof}{\texttt{cutoff time}\xspace}
\usepackage{tikz,pgfplots,sansmath}
\usetikzlibrary{patterns}

\begin{document}

\title{MLFriend: Interactive Prediction Task Recommendation for Event-Driven Time-Series Data}

\author{
Lei Xu, Shubhra Kanti Karmaker Santu and Kalyan Veeramachaneni\\
Laboratory for Information \& Decision Systems\\
Massachusetts Institute of Technology\\
Cambridge, MA, 02139, USA\\
\texttt{\{leix,santu,kalyanv\}@mit.edu}
}


\maketitle

\begin{abstract}
\begin{quote}
Most automation in machine learning focuses on model selection and hyper parameter tuning, and many overlook the challenge of automatically defining predictive tasks. We still heavily rely on human experts to define prediction tasks, and generate labels by aggregating raw data. In this paper, we tackle the challenge of defining useful prediction problems on event-driven time-series data. We introduce \mlf to address this challenge. \mlf first generates all possible prediction tasks under a predefined space, then interacts with a data scientist to learn the context of the data and recommend good prediction tasks from all the tasks in the space. We evaluate our system on three different datasets and generate a total of 2885 prediction tasks and solve them. Out of these 722 were deemed useful by expert data scientists. We also show that an automatic prediction task discovery system is able to identify top 10 tasks that a user may like within a batch of 100 tasks.
\end{quote}
\end{abstract}

\input{1.intro.tex}

\input{2.task.tex}
\input{5.related.tex}

\input{3.model.tex}
\input{4.exp.tex}

\input{6.conclusion.tex}

\bibliographystyle{aaai}
\bibliography{sample-bibliography}

\end{document}

%% file: 1.intro.tex
\section{Introduction}
A primary goal of machine learning over event-driven time series datasets is to learn patterns and predict future outcomes. However, when faced with such large interconnected data, even experts find it difficult to decide what predictive tasks to focus on. We investigated the $15$ most popular event-driven time series datasets on Kaggle for which no explicit prediction goal was defined. Only $34\%$ of kernels\footnote{A kernel is a piece of code uploaded by a Kaggle user trying to process data. For this survey, we checked $10$ most popular kernels for each dataset.} were focused on building predictive models\footnote{We define making a predictive model as training a model to predict a future outcome.}, while the majority simply helped to show statistics and visualize data. If we ignore datasets that pertain to stock price and product price where predicting the price column is an obvious prediction task, the percentage of kernels exploring predictive models drops to $23\%$.


\begin{figure}[t]
    \centering
    \includegraphics[width=\columnwidth]{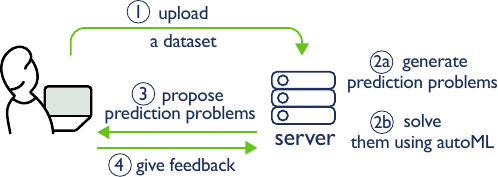}
    \caption{The procedure of using \mlf to discover useful prediction tasks for a new dataset. \mlf shown here as server serves proposes prediction tasks to the user/data scientist, who can give feedback and ranking. }
    \label{fig:overview}
\end{figure}

Even for data science enthusiasts, defining prediction tasks on event-driven data is challenging, especially when the most interesting column is not very obvious. This triggers an interesting research question: Is it possible to build a system that can formulate useful prediction tasks automatically? Formulating a prediction task precisely is non-trivial. Natural language questions are vague and usually are not grounded in available data, and data scientists need to specify a lot of details and perform a lot of pre-processing before they can even begin doing machine learning. 

Take for example a so called machine learning goal expressed as: predicting the likelihood of future flight delays. First, the user must specify the entity in question. In the case above, we are not sure whether we need to make predictions for each flight number, each aircraft, or each airline company. Second, they need to figure out how to quantify the outcome they want to predict. For example, they can build a classification model to predict whether a particular flight will be delayed or not, or they can build a regression model to predict how many flights will be delayed and the average delay time. Finally, the prediction \textit{window} is also unknown. For example, the user may want to predict flight delays in the next few hours or as far out as next year. By specifying these details, a simple sentence can become tens of prediction tasks, all with different levels of usefulness and perhaps with different prediction accuracies.

For a computer to generate these tasks and process the data, this formulation requires a precise mathematical representation. But to the best of our knowledge, no such representation currently exists for prediction tasks. To achieve this, in this paper we represent a prediction task using a sequence of operations over a dataset pertaining to a time period. We came up with these operations by surveying a large repository of prediction tasks. The challenge of however lies in balancing coverage with usefulness.  Thus we limit these operations to a small subset of what is possible but extensible. 



Given a general expression for representing a prediction task, computers can quickly enumerate over all possible instantiations of that representation to generate thousands of prediction tasks. For example, predicting the likelihood of future flight delays can be enumerated as predicting the number of flights delayed by more than 5 minutes for a particular airline, predicting the number of flight delays at a particular airport, predicting the average delay of all flights run by a particular airline and several others. With a large number of tasks generated, another natural challenge is how to discover the most \textit{useful} ones. To this end, we propose an interactive recommendation method. We train a model to learn the context of the data and the user's need during this interaction. We call this the ``\textit{automated task discovery}'' problem (ATD). 

Fig. \ref{fig:overview} shows how data scientists use \mlf to discover interesting prediction tasks on a new dataset. Once they plug in a new dataset, the system will show $k$ prediction tasks, then the data scientist will provide feedback by marking some tasks as \textit{important}, \textit{interesting} or \textit{useful} per his/her preferences. The system then uses the data scientist's feedback to learn more about the user preferences and thus, provides better recommendations in the next iteration. To validate our system, we annotate prediction tasks on 3 datasets, and show that our interactive recommendation system can make personalized recommendations effectively.

\textbf{Our contributions through this paper are as follows}:
\begin{itemize}
    \item[--] To the best of our knowledge, we are the first to formulate a way to represent a prediction task space that is tractable, \textit{useful} and extendable.
    \item[--] We define ATD, a new recommendation problem over prediction tasks and then, propose an effective interactive algorithm for recommending useful tasks. 
    \item[--] We develop an end-to-end system which automatically generates prediction tasks, extracts training examples, makes training and validation sets, and applies automatic machine learning methods (feature engineering, model selection and tuning) and presents the results. To achieve this we stand on the shoulders of researchers who have developed and packaged several AutoML solutions.
\end{itemize}

The paper is organized as follows. Section~\ref{sec:overview} presents a brief overview of the \mlf system. Section~\ref{related} presents the related work. Section~\ref{apg} presents our approach for generating prediction tasks automatically. Section~\ref{sec:rec} describes the interactive recommendation method. Section~\ref{settings} presents experimental results and Section~\ref{sec:disc} provides discussion of our results. Section~\ref{concludes} gives our conclusion and future work.

%% file: 2.task.tex
\section{MLFriend Overview}\label{sec:overview}
To assist data scientists in formulating and solving prediction problems, we need to build a system that can automatically generate prediction tasks, and help them find the most useful task to solve. Our system, dubbed \texttt{MLFriend}, fulfills these goals. We first present a brief overview of the type of data \mlf can operate on and subsequently,  describe the overall architecture of \mlf.

\noindent {\bf Event-Driven Time Series Data: } Our focus in this paper is event-driven time series tables. Event-driven time series data takes the form of an incremental table, where an additional row is added each time an event happens. Each row includes a \texttt{timestamp} when the event happened, related entities, and other properties or attributes of the event. Event-driven data are very common; e.g. flight records, online purchase records, and online education records all take this form.

We formally define event-driven time series data as a table $T$ containing $N_e+N_d+1$ columns, say $\{\mathbf{t}, \mathbf{e}_1, \mathbf{e}_2, \ldots, \mathbf{e}_{N_e}, \mathbf{d}_1, \mathbf{d}_2, \ldots, $ $\mathbf{d}_{N_d}\}$, where $\mathbf{t}$ is the time column, $\mathbf{e}_i$ is an entity column, and $\mathbf{d}_i$ is another numerical or categorical column. In our example, $\mathbf{t}$ is the date-time column, $\mathbf{e}_1$ and $\mathbf{e}_2$ are the flight number column and the airline column respectively, and $\mathbf{d}_1$ is the flight-delay indicator column. For simplicity, we only consider data with one time column. We select an entity column from all entity columns $\mathbf{e}^* \in \{\Phi, \mathbf{e}_1,\ldots \mathbf{e}_{N_e}\}$ and our goal is to make prediction tasks for each entity $e\in\mathbf{e}^*$. For example, if we choose the airline column as $\mathbf{e}^*$, we will then create tasks to make predictions for each airline. When $\mathbf{e}^*=\Phi$, we consider all records to be related to a pseudo root entity, so that we create prediction tasks that attempt to make predictions for all the records. 

\begin{figure*}[t]
    \centering
    \includegraphics[width=0.85\textwidth]{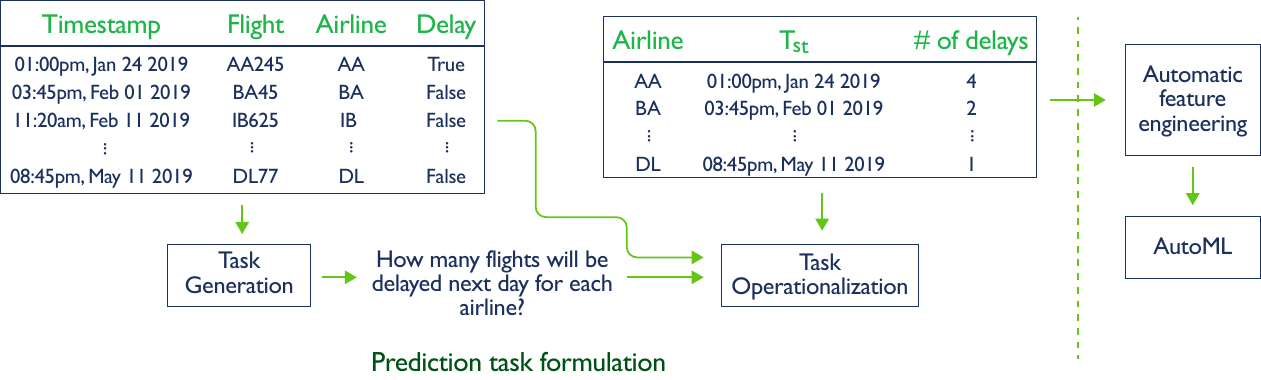}
    \caption{An example dataset and the process of generating prediction tasks and solving them. We start with a raw event-driven time series data. We generate prediction tasks using the prediction task generator. Here, we generate \textit{How many flights will be delayed the next day for each airline?}. The task operationalization box identifies the training examples from the historic raw data and outputs a list of training examples shown in the table above it. This table contains the list of training examples, with the entity for which we are making a prediction, the time point at which the prediction is sought and the outcome that is to be prediction. This table is then provided to automatic feature engineering (AFE) tools and whose output is then sent to \texttt{AutoML} systems to develop a machine learning model. For each row in the table, the automated feature engineering tools raw data before $t_{st}$ to generate a feature vector. We match features and labels to create instances for machine learning models. Many tools are available For AFE and AutoML. In this paper the focus is on the prediction task formulation. 
    }
    \label{fig:pipeline}
\end{figure*}

\noindent {\bf Architecture Overview:} Fig.\ref{fig:pipeline} illustrates the entire workflow, using the example of a flight delay table with only 4 columns: a datetime column, a flight number column, an airline company column, and a boolean column showing whether or not the flight was delayed. To start the process, a user uploads this dataset to our system, and specifies the data type for each column. \mlf responds to the user by performing two primary functions as mentioned below:

\begin{enumerate}
    \item{\bf Prediction Task Generation:} The system asks for the entity column for which the user is seeking to develop predictive models, as well as the prediction window size. Based on the user response, the system generates hundreds of  Prediction Tasks for this entity using a Prediction Task Representation Language (described in section~\ref{PTG}). An example task for the entity ``airline'' might be: \emph{For each airline, predict how many flights will be delayed the next day?}
    
    \item {\bf Prediction task operationalization:} Once a prediction task is formulated, the next important job for \mlf is to extract training examples from the historical data. To achieve this, \mlf generates several time points, as well as target labels at each time point. Now, let us consider the same prediction task again, i.e., \emph{For each airline, predict how many flights will be delayed the next day?} To generate the target value for this prediction task, for each airline, the system applies two operations to the data after each time point: (1) an equal filter operation and (2) a count aggregation operation. In this way, \mlf generates the target value at each of the cutoff times.
\end{enumerate}

With the two functions, \mlf has created a prediction task and extracted training examples. At this point, the user can decide to employ machine learning to solve the task and see accuracy, or request \mlf to generate more tasks. The user can then rate these tasks as s/he interacts with the \mlf. 

\noindent \textbf{Solving the prediction task}: \mlf utilizes the \texttt{AFE} (automated feature engineering) and \texttt{AutoML} frameworks available to develop a machine learning model for the prediction task. 

%% file: 5.related.tex
\section{Related Work}\label{related}
Applying machine learning to a practical task requires several steps, including task definition, data preparation, feature engineering, and model and hyperparameter selection. Because each step requires human involvement and expert knowledge, this process is time-consuming, and the machine learning community spent a lot of effort designing algorithms and building software to automate one or more of its aspects. \cite{katz2016explorekit,kanter2015deep,mountantonakis2017linked,van2017automatic,khurana2017feature,kaul2017autolearn} can all extract useful features from data and generate feature matrices. Simple machine learning methods, including nearest neighbor searches, decision trees, and support vector machines, are sensitive to input features. (Neural networks depend less on features because they can autonomously learn a task-specific vector, but they are more sensitive to network structure and hyperparameters.) Random search and sequential modeling methods \cite{bergstra2012random,snoek2012practical,hutter2011sequential,bergstra2011algorithms,bengio2012practical,bergstra2013making,swersky2013multi,maclaurin2015gradient} are used in hyperparameter tuning. \cite{zoph2016neural,zoph2017learning,liu2017hierarchical,liu2017progressive,real2018regularized,pham2018efficient,baker2017accelerating} propose neural or reinforcement learning models to search optimal neural network structures. \cite{thornton2013auto,feurer2015efficient,thornton2013auto,swearingen2017atm} are wrappers for existing machine learning libraries, providing efficient model selection and hyperparameter tuning functions. 

These models and software have enabled automation from the classification or regression task to the prediction result. However, to the best of our knowledge, no one has tried to automate the first step in the machine learning process: defining prediction tasks. This missing piece prevents the completion of a fully automated pipeline that transforms a dataset into a useful predictive model and garners results. 




%% file: 3.model.tex
\section{Automatic Prediction Task Generation}\label{apg}
In this section, we propose a representation for prediction tasks and propose the automatic generation of hundreds of prediction tasks. 

\subsection{Prediction Task Generation}\label{PTG}
The Prediction Task Generation process can be divided into two distinct steps, namely, Prediction Task Representation and Prediction Task Enumeration. Below, we present these steps.

\noindent {\bf Prediction Task Representation:} The goal of any prediction task is to estimate a target label for an entity at a certain time point. Thus, the goal generated by a prediction task can be defined as a three tuple $<e_i, \mathtt{t_s}, label>$. For an event-driven time series, a target label is a result of a sequence of atomic operations applied to the data \textit{after} the $\mathtt{t_s}$. Each operation can (1) remove some rows, (2) add a new column to the data, or (3) compute some aggregation. There are two major categories of operations: filter operations and aggregate operations. Filter operations filter out noisy rows and keep rows with some particular property. $\texttt{fil\_op}$, $\epsilon$ and $\mathbf{d}_f \in \{None, \mathbf{e}_{1:N_e}, \mathbf{d}_{1:N_d}\}$ are the filter operation, its hyperparameter and corresponding filter column. $\texttt{fil\_op}_(T, \mathbf{d}_f, \epsilon)$ returns some rows of $T$ with column $\mathbf{d}_f$ satisfying some property.

Aggregation operations aggregate multiple rows and generate a target label. \texttt{agg\_op} and $\mathbf{d}_g \in \{None, \mathbf{e}_{1:N_e}, \mathbf{d}_{1:N_d}\}$ are the aggregation operation and corresponding aggregation column respectively. $\texttt{agg\_op}(T, \mathbf{d}_g)$ returns a numerical or categorical value using data in column $\mathbf{d}_g$. Similar to SQL generation \cite{zhongSeq2SQL2017}, we pick a predefined filter operation set $\mathbf{O}_f$ and predefined aggregation operation set $\mathbf{O}_g$ as shown on Table \ref{tab:ops}.

For the example task \emph{for each airline, predict how many flights will be delayed the next day}, we use an equal filter operation to find all the delayed flights and then use a count aggregation operation to count the number of delayed flights for each airline company. Filter operations usually need a hyperparameter. We further constrain prediction tasks to only 1 filter operation and 1 aggregation operation. A sequence of 2 operations gives us a tractable space, while incorporating more operations can increase the possibilities. 

We now formally define a executable prediction task on a event driven time series table $T$. An executable prediction task is defined as a 7-tuple $(\mathbf{e}^*, \texttt{fil\_op}, \epsilon, \mathbf{d}_f, \texttt{agg\_op}, \mathbf{d}_g, \mathbf{t}_s, \mathbf{p}_w)$. $\mathbf{e}^*$ is the entity column for the task, $\mathbf{t}_s$ is the prediction start time and $\mathbf{p}_w$ is the prediction window. We further define a prediction task template as a 5-tuple:  $(\mathbf{e}^*, \texttt{fil\_op}, \mathbf{d}_f, \texttt{agg\_op}, \mathbf{d}_g)$. It's very similar to an executable prediction task except that we generalize it by not including the hyperparameter $\epsilon$, the start time $\mathbf{t}_s$ and prediction window $\mathbf{p}_w$.

\begin{table}[htb!]
    \centering
    \small
    \setlength{\tabcolsep}{0.5\tabcolsep}
    \begin{tabular*}{\linewidth}{@{\extracolsep{\fill}}lcc@{}}
        \toprule
        Operation Set & Supported Ops & Supported Types \\\midrule
        Filter & \texttt{all\_fil} & None\\
        $\mathbf{O}_f$    & \texttt{greater\_fil}, \texttt{less\_fil} & Numerical\\
        & \texttt{eq\_fil}, \texttt{neq\_fil} & Categorical/Entity\\\midrule
        Aggregation & \texttt{count\_agg} & None\\
        $\mathbf{O}_g$    & \texttt{sum\_agg}, \texttt{avg\_agg}, & Numerical \\
           & \texttt{min\_agg}, \texttt{max\_agg} & Numerical \\
        & \texttt{majority\_agg} & Categorical/Entity \\
        \bottomrule
    \end{tabular*}
    \caption{Predefined filter and aggregation operation sets.}
    \label{tab:ops}
\end{table}

\begin{algorithm}[htb]
\SetAlgoLined
\KwIn{Entity Column $\mathbf{e}^*$, Table $T$.}
\KwResult{A list of prediction task templates.}
\If {$type(\mathbf{e}^*) \notin \{\Phi, Entity, Categorical\}$}{
    \Return $[]$\;
}
$result \leftarrow []$\;
\For{
$\texttt{fil\_op} \in \mathbf{O}_f$, 
$\texttt{agg\_op} \in \mathbf{O}_g$, 
$\mathbf{d}_f \in \{None, \mathbf{e}_{1:N_e}, \mathbf{d}_{1:N_d}\}$, 
$\mathbf{d}_g \in \{None, \mathbf{e}_{1:N_e}, \mathbf{d}_{1:N_d}\}$}{
    \If {$type(\mathbf{d}_f) \in supported\_types(\texttt{fil\_op})$ and
        $type(\mathbf{d}_g) \in supported\_types(\texttt{agg\_op})$} {
        $result.append\big((\mathbf{e}, \texttt{fil\_op}, \mathbf{d}_f, \texttt{agg\_op}, \mathbf{d}_g)\big)$\;
    }
}
\Return $result$\;
\caption{Prediction task enumeration}
\label{alg:gen}
\end{algorithm}


{\bf Prediction Task Enumeration:} Using our representation of prediction tasks, the generation of these tasks is intuitive. As shown in Algorithm \ref{alg:gen}, we exhaustively create all possible combinations of operations and columns, then check whether the data type of such combinations is correct. Since the algorithm does not execute operations on data, it only needs the data schema. The complexity depends on the size of the operation set and the number of columns, both of which are usually small. The upper bound of prediction tasks using predefined operation sets in Table \ref{tab:ops} is
\begin{equation}
    \big(2(N_c+N_e-1)+2N_n+1\big)\times\big(N_c+N_e-1+4N_n+1\big),
\end{equation}
where $N_e$, $N_c$, $N_n$ are the number of entity columns, categorical columns and numerical columns respectively ($N_c+N_n=N_d$). 
Users can try all possible entities by changing $e^*$.



\subsection{Prediction task operationalization}
\label{sec:cutoff}
Given a prediction task, operationalization refers to creating training examples for machine learning. This involves two steps: (1) setting the values for the hyperparameters on the prediction task template, and (2) identifying the training examples from the historical data. 

\noindent \textbf{Setting the hyperparameters}:  To finalize the task, one needs to set the hyperparameters for the template. These are any $\epsilon$ that the operations require and the prediction window, $p_w$. In our experience these are usually application-specific hyperparameters that experts prefer to set. The choice of hyperparameter may not affect the meaningfulness and usefulness of the task, but may affect its practical application. For example, a user may prefer to predict ``\textit{number of delays}'' for the next week instead of the next day. In many cases, experts prefer to vary the hyperparameters to see how they affect the predictability. 

\noindent \textbf{Identifying training examples}: Identifying training examples requires us to scan the historical data of each entity and identify tuples of the form $<e \in\mathbf{e}^*, t_s, label>$. We use a fixed-window repeating time to specify multiple $t_s$. With $\mathbf{p}_w$ defined, for each $e \in \mathbf{e}^{*}$ we define a base time point $t_{base}$ and the terminate time point $t_{terminate}$. We fit $k$ back-to-back windows between $t_{base}$ and $t_{terminate}$ as $\{(t_{st}^{(1)}, t_{ed}^{(1)}), \ldots, (t_{st}^{(k)}, t_{ed}^{(k)})\}$, where $t_{st}^{(i + 1)}=t_{ed}^{(i)}$ and $t_{st}^{(i)}+\mathbf{p}_w = t_{ed}^{(i)}$. Then each entity from the selected entity column $e\in\mathbf{e}^*$ can be combined with $k$ windows as 3-tuples $(e, t_{st}^{(i)}, t_{ed}^{(i)})$. We can clearly write the start and end time of each prediction window, $\mathbf{p}_w$, in a table as shown in Table \ref{tab:egcutoff}. Under this representation, label generation is straightforward. We generate one label for each row $(e, t_{st}, t_{ed})$ in the table. We extract all related rows from T as 
\begin{equation}
T'=\{r| r\in T, r[\mathbf{e}^*]=e, t_{st}\leq r[\mathbf{t}]<t_{ed}\}.
\end{equation}
Then we apply the filter operation and aggregation operation successfully to generate the label as
\begin{equation}
    l=\texttt{agg\_op}\big(\texttt{fil\_op}(T', \mathbf{d}_f, \epsilon), \mathbf{d}_g\big).
\end{equation}

\noindent {\bf The concept of Cutoff Time:} Machine learning model development for prediction tasks, using training examples generated from historical data as described above, usually requires rolling back in time and emulating the conditions at each of the $t_s$. The system should be carefully designed so that data used to generate features doesn't leak the label. To accurately do this, we introduce the concept of \cof, and describe how cutoff time regulates the usage of the data.   We use the data before $t_{st}$ to generate features, and we use data between $t_{st}$ and $t_{ed}$ to generate labels. Thus $t_{st}$ represents \cof. Current sophisticated automated feature engineering tools accept \cof{s} as input along side the data \footnote{Featuretools: https://docs.featuretools.com/}. 

\begin{table}[htb]
    \centering
    \begin{tabular*}{\linewidth}{@{\extracolsep{\fill}}lcc}
        \toprule
        Entity & $t_{st}$ & $t_{ed}$ \\\midrule
        AA & Jan. 1 & Jan. 8 \\
        AA & Jan. 8 & Jan. 15 \\
        AA & Jan. 15 & Jan. 22 \\
        AA & Jan. 22 & Jan. 29 \\\midrule
        DL & ... & ...  \\
        \bottomrule
    \end{tabular*}
    \caption{Fixed-window cutoff time table example. We choose the airline as target entity column. $t_{base}$ is Jan. 1. $t_{terminate}$ is Jan. 31.  $\mathbf{p}_w$ is 7 days. $t^*$ is Jan. 15. }
    \label{tab:egcutoff}
\end{table}

\section{Learning to Recommend Prediction Tasks}
\label{sec:rec}

Automated task discovery (ATD) involves discovering interesting prediction tasks within our task space. ATD needs to be personalized because the usefulness of a task depends on the user's high-level goal. For example, airline companies are more interested in which airport will have an unusual amount of delayed flights so that they can accommodate passengers, while passengers are more interested in whether their flights will be delayed. Hence, a useful ATD system should heuristically adapt to user's preference. 

We formally define ATD as follows:  Given a dataset $T$ and a user defined parameter $n$, ATD generates $n$ different valid prediction tasks $X=\{x_1, \ldots, x_n\}$. ATD ranks these prediction tasks using a ``\textit{meta model}'' \footnote{We call this meta model to distinguish between it and the model built to solve the prediction task itself. We remind that our goal is to use machine learning itself to rank machine learning tasks} and recommends highly ranked tasks to the user one by one or in a batch fashion. The user then provides feedback by marking each recommended task as useful or not useful. ATD then uses this feedback to revise the ``\textit{meta model}'' to better accommodate the user's preference and re-ranks the prediction tasks accordingly. This revision process can continue in an iterative fashion until the user is satisfied with the recommendations.


At first, this set up might suggest that collaborative filtering could be used for the so called ``\textit{meta model}''. But when applied to ATD, collaborative filtering methods suffer from two problems:

\begin{itemize}
\item Cold start: Nowadays, we have lots of different event-driven data and some of them are private. A typical scenario would involve only one data scientist working on a dataset. In this case, collaborative filtering does not work at all for that user and that dataset. 
\item Lack of training data for the ``\textit{meta model}: Typical recommendation systems, such as movie recommenders, are trained on thousands of items and millions of users. In ATD, getting that much data is challenging. Our system targets expert data scientists, who are a much smaller population, and whose time is very expensive, which makes this data availability even more sparse.
\end{itemize}

For these reasons, instead of learning correlations between different users and different data sets, our algorithm iteratively learns a new \textit{meta model} for each dataset and each user.

\subsection{Feature Representation for Task Recommendation}
Any recommender system needs a general feature representation of the entity it aims to recommend. In our case, the feature vector is composed of (1) one-hot representations for the entity column, filter operation, filter column, aggregation operation and aggregation column, and (2) the estimated goodness of each attribute, i.e the proportion of good prediction tasks among all tasks with that column or operation - so far. The one-hot representations and goodness scores of all attributes and operations of a task are concatenated into a single vector, which is then used as the feature representation of that particular task.

\subsection{Task Discovery Model}
In each iteration, the system proposes several tasks, and the user marks the best ones. The system then learns from user's feedback and makes a better recommendation in next iteration. We use the score prediction model for recommendation. Due to small size of training set, we use linear regression to predict the score. In each iteration $r$, the system proposes tasks $\{x_r^{(1)}, $ $\ldots, x_r^{(k)}\}$. A data scientist will then give feedback on each task as $\{y_r^{(1)}, \ldots, y_r^{(k)}\}$. In the next iteration $r+1$, we learn a new linear model with objective
\begin{equation}
\mathcal{L} = \sum_{i=1}^{r} \sum_{j=1}^{k} \big(f(x_i^{(j)})\cdot \theta - (y_i^{(j)})^2\big) + \alpha |\theta|^2,
\end{equation}
where $f(\cdot)$ converts a prediction task to a real-valued feature vector and $\alpha$ is a regularization hyperparameter and $\theta$ are the parameters for the linear model.  The optimum model $\theta^*=\arg_\theta\min \mathcal{L}$, then greedily selects the top $k$ tasks with the highest $f(x)\cdot\theta^*$ in the remaining tasks. Algorithm \ref{alg:pd} shows the recommendation in $r+1$ iteration given all the feedback from previous $r$ iterations. 

\begin{algorithm}[htb]\small
\SetAlgoLined
\KwIn{\\
All tasks: $S$\\ 
Previous Shown tasks: $X=\{x_{1:r}^{(1:k)}\}$\\
Previous Feedbacks $Y=\{y_{1:r}^{(1:k)}\}.$\\}
\KwResult{A list of $k$ tasks.}\BlankLine\BlankLine\BlankLine

$\theta^*_{r+1} \leftarrow \arg_\theta\min\sum\limits_{i=1}^{r} \sum\limits_{j=1}^{k} \big(f(x_i^{(j)})\cdot \theta - (y_i^{(j)})^2\big) + \alpha |\theta|^2$\;
$task\_score \leftarrow []$\;
\For {task $\in S\setminus X$} {
    $task\_score.append((task, featurize(task)\cdot \theta^*_{r+1}))$\;
}

$x_{r+1}^{(1:k)} \gets top\_k(task\_score)$\;
$y_{r+1}^{(1:k)} \gets gather\_feedback(top\_k(task\_score)$\;
$X\gets X\cup x_{r+1}^{(1:k)}$\;
$Y\gets Y\cup y_{r+1}^{(1:k)}$\;

\Return $top\_k(task\_score)$

\BlankLine\BlankLine\BlankLine\BlankLine
\SetKwProg{Fn}{Function}{}{}
\Fn{featurize(task)}{
    $feature \leftarrow get\_feature(task, \text{`entity\_col'})$\;
    $feature \leftarrow feature \oplus get\_feature(task, \text{`fil\_col'})$\;
    $feature \leftarrow feature \oplus get\_feature(task, \text{`agg\_col'})$\;
    $feature \leftarrow feature \oplus get\_feature(task, \text{`fil\_op'})$\;
    $feature \leftarrow feature \oplus get\_feature(task, \text{`agg\_op'}')$\;
    \Return $feature$\;
}

\BlankLine\BlankLine

\Fn{get\_feature(task, attribute)}{
    $n\_good \leftarrow 0$\;
    $n \leftarrow 0$\;
    \For {$(x, y) \in \langle X, Y\rangle$}{
        \If{task.attribute = x.attribute}{
            $n \leftarrow n + 1$\;
            \If {y = 1} {
                $n\_good \leftarrow n\_good + 1$\;
            }
        }
    }
    $goodness\gets\frac{(n\_good+1)}{(n+1)}$\;
    \Return $one\_hot(task.attribute)\oplus goodness $
}

\caption{Task Discovery}
\label{alg:pd}
\end{algorithm}

%% file: 4.exp.tex
\section{Experimental Settings and Results}\label{settings}
\noindent \textbf{Datasets}: We pick $3$ event-driven time series datasets from Kaggle and generate prediction tasks on them. \textbf{Chicago Bicycle}\footnote{https://www.kaggle.com/yingwurenjian/chicago-divvy-bicycle-sharing-data} records the use of shared bicycles in Chicago between 2014 and 2017. \textbf{Flight Delay}\footnote{https://www.kaggle.com/usdot/flight-delays} is the 2015 flight records in US. \textbf{YouTube Trending}\footnote{https://www.kaggle.com/datasnaek/youtube-new} records trending YouTube videos between 2017 and 2018. Each dataset has a time column. We pick the time column, entity columns, numerical columns, and categorical columns, while removing unsupported columns like sets or natural language. Table \ref{tab:ncol} shows the number of columns we use to generate prediction tasks in each dataset. 

\input{datasetdetail.tex}

\subsection{Task Generation}
For each dataset, we pick several entity columns and run prediction task generation algorithms on these entities. Table \ref{tab:exp} shows the number of prediction task templates we can generate on each dataset and each entity. 

The number of prediction task templates depends on the number and type of columns. The flight delay dataset has 19 columns -- more than other datasets -- so we generate the most prediction tasks on that dataset. Even though it is a big dataset, we generate 1680 tasks, which is better than infinite possibilities. For other datasets, we generate a few hundred prediction tasks for each entity. 

Since we have the same number of filter operations for each type, but more aggregation operations for numerical columns than categorical columns in the predefined operation set, the number of prediction tasks is more sensitive to the number of numerical columns. Our system also generates more regression tasks than classification tasks, simply because only the majority aggregation operation generates classification tasks while all other aggregation operations generate regression tasks.

\begin{table}[htb]
\begin{tabular*}{\linewidth}{@{\extracolsep{\fill}}lccc}
\toprule
Dataset and Entity & \#Reg & \#Cls & \#Tot   \\
\midrule
Chicago Bicycle    & 320/612       & 99/163            & 419/775   \\
\hspace{1em}--from station id   &    211/289    &   55/68    & 266/357   \\
\hspace{1em}--$\Phi$              &     109/323      &     44/95      & 153/418   \\\midrule
Flight Delay       & 1757/2870     & 318/490           & 2075/3360 \\
\hspace{1em}--airline           &   898/1435   &  133/245        & 1031/1680 \\
\hspace{1em}--origin airport    &   859/1435  &  185/245         & 1044/1680 \\\midrule
YouTube            & 353/595       & 38/48             & 391/643   \\
\hspace{1em}--category id       &     153/187          &        9/11           & 162/198   \\
\hspace{1em}--channel title     &      162/187         &         10/11          & 172/198   \\
\hspace{1em}--$\Phi$              &        38/221       &         19/26          & 57/247    \\\bottomrule
\end{tabular*}
\caption{Prediction task templates on each dataset. For each dataset, we choose several different entities for task generation as listed in first column. Each entry in the table shows \#valid/\#all. \#all means the number templates with correct data types. \#valid means that at least one hyperparameter can make that template find a training set with at least 10 instances and a validation set with at least 5 instances. }
\label{tab:exp}
\end{table}

\subsection{Building machine learning models}
After we generate prediction tasks, we develop machine learning solutions using available automated feature engineering tools and \texttt{AutoML} tools. 
 
 \noindent \textbf{Set task hyperparameters}: To accomplish this we first set the hyperparameters for each prediction template. 
\begin{itemize}
    \item For filter operations on numerical columns,  we select a threshold which keeps a ratio of data. Typically, we set target ratio to be $25\%, 50\%, 75\%$. We sample some rows from $T$, then we choose a threshold which keeps the ratio of data as close to our target as possible. 
    \item For operations on categorical columns, we try top $3$ majority categories. 
\end{itemize}

\noindent \textbf{Train test splits} We pick a time $t^*$ to split the training and validation sets. We use all the data, with $t_{st} < t^*$ as training data and the other data as validation data. When compared with cross-validation, our splitting strategy has the following advantages:
\begin{itemize}
    \item It prevents feature leakage. If allowed to randomly choose training and validation sets, the machine learning model can "cheat" by remembering some future information from training examples. For example, if Jan. 1 has an unusually high probability of flight delay, a machine learning model can "see" from the training set that a flight on this date is more likely to be delayed, and predict such for the validation set. Although this information helps the model to perform well on the validation set, it is impossible to replicate this performance on a real test set, because this future information is not available--when it is making predictions about Jan. 2, such information does not exist. 
    \item  It is more similar to the actual application. In a real production environment, people update the model regularly when they collect new data, so each model is only responsible for the next few prediction windows. Here we validate the model by predicting the most recent windows. Because this mimics a real circumstance, the validation result closely approximates the model's performance on a real application.  
\end{itemize}

\noindent \textbf{Feature engineering and modeling}: To build the model, we use an automated feature engineering package \texttt{Featuretools}\footnote{https://docs.featuretools.com/} to generate features for each training example. We use \texttt{scikit-learn} and grid search to build, tune and test a machine learning model.


\begin{figure*}[htb!]
   \centering
     \includegraphics[width=.85\textwidth]{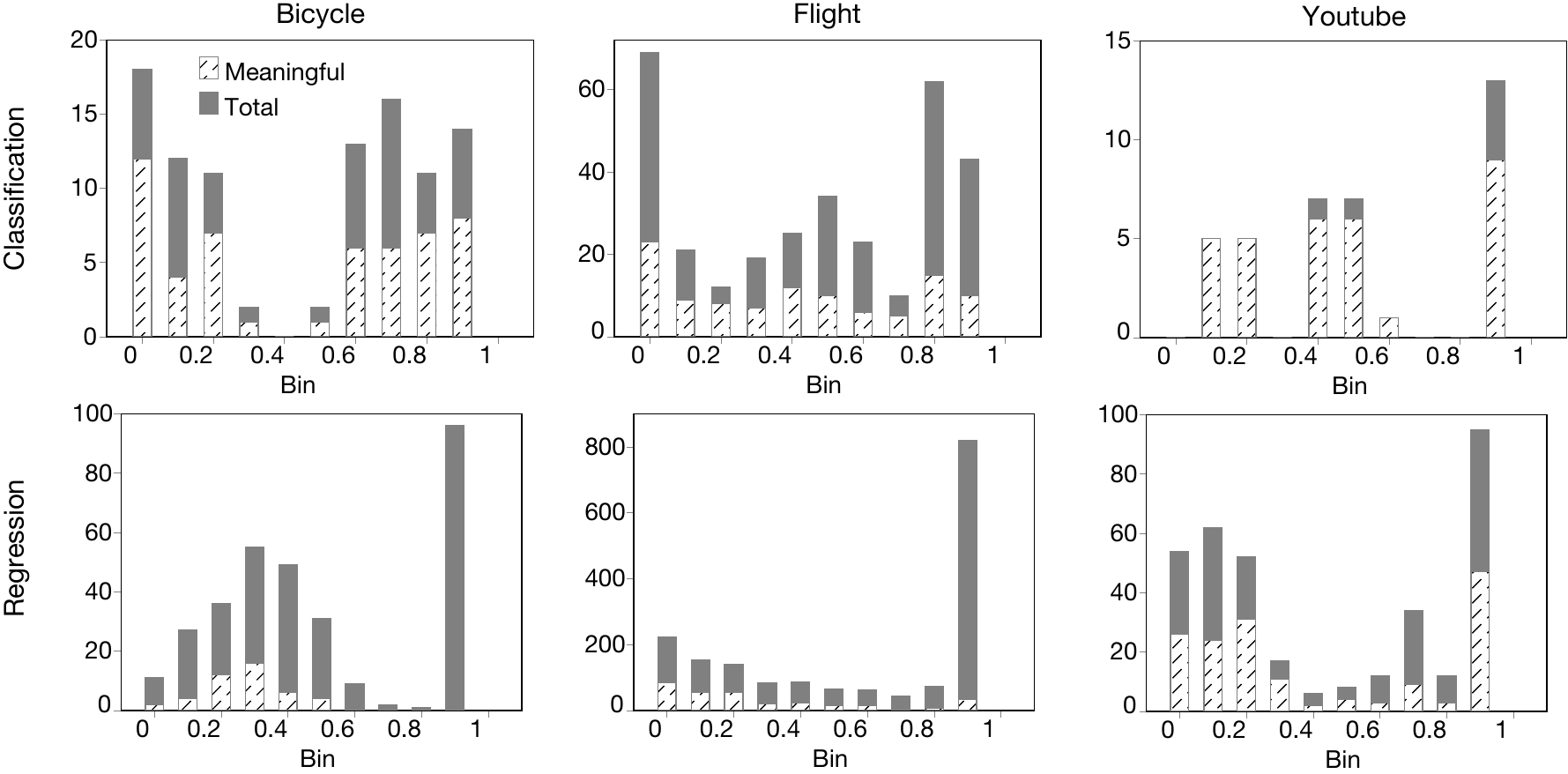}
   \includegraphics[width=.9\columnwidth]{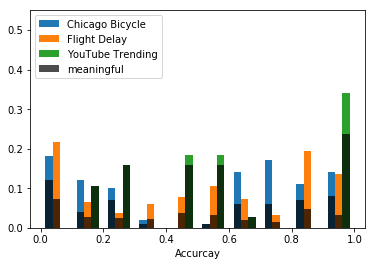}
   \caption{Top: Distribution of accuracies for classification tasks on three datasets. Bottom: Distribution of $R^2$ for regression tasks on three datasets.}
     \label{fig:hist}
 \end{figure*}

\subsection{Creating Ground Truth for Evaluation}
\label{sec:ann}
To evaluate the value proposition of our task generation system and learning efficacy of our APD system, we needed to identify the most interesting prediction tasks in the task space. Hence we decided to manually annotate the tasks, and then identify if the algorithm for APD as described in Section~\ref{sec:rec} can identify the most preferred tasks automatically. We built a web interface to annotate prediction tasks generated by our system, and posted our annotation task on Mechanical Turk. However, the workers there couldn't understand prediction tasks easily, and after getting random annotations, we decided to annotate the data ourselves. $9$ expert data scientists helped compare prediction tasks while $2$ of them finished most of the annotations. 

\begin{table}[htb]
    \centering
    \setlength{\tabcolsep}{0.5\tabcolsep}
    \begin{tabular*}{\linewidth}{@{\extracolsep{\fill}}lcccc}\toprule
        Dataset & \#Valid &\#Meaningless & \#Cmp & \#Per\\\midrule
        Chicago Bicycle & 419 & 320 & 200 & 4.04 \\
        Flight Delay & 2075 & 1660 & 416 & 2.00\\
        YouTube Trending & 391 & 183 & 350 & 3.37 \\\bottomrule
    \end{tabular*}
    \caption{Annotation statistics. \#Valid means the number of valid prediction tasks. \#Meaningless the number of meaningless tasks marked by annotators and \#Cmp is the number of comparison between meaningful tasks respectively. \#Per means the average number of comparison each meaningful task have where $\text{\#Per}=2\times\text{\#Cmp}/(\text{\#Valid}-\text{\#Meaningless})$. }
    \label{tab:ann}
\end{table}

\noindent \textbf{Annotation Settings}:  Simply rating each task from 1 to 5 doesn't work because different annotators have different scoring means and deviations, and the scoring criterion changes during the annotation for a single annotator. Recent progress in crowdsourcing data collection suggests comparison based ranking \cite{jamieson2011active,wauthier2013efficient}. We use the win-lose-tie scoring to compare tasks. Each pair of prediction tasks $(A, B)$ is compared on two metrics: meaningfulness and usefulness. 
\begin{itemize}
    \item Meaningfulness (m) means that people can understand the prediction task.
    \item Usefulness (u) means that the prediction is useful in making some high-level decision. 
\end{itemize}
In each comparison, $A$ can win, lose or tie with $B$ in each metric. We define the scoring function for each metric as 
\begin{equation}
    S_{metric}(A, B) =
    \begin{cases}
        3, & \text{ if $A$ is better than $B$},\\
        1, & \text{ if $A$ is as good as or as bad as $B$},\\
        0, & \text{ if $A$ is worse than $B$}.
    \end{cases}
\end{equation}
We weigh the metrics as $w_{m}=0.7$ and $w_{u}=0.3$. So, in each comparison, the score A and B get is 
\begin{align}
    s_A &= w_{m} * S_{m}(A, B) + w_{u} * S_{u}(A, B).\\
    s_B &= w_{m} * S_{m}(B, A) + w_{u} * S_{u}(B, A). 
\end{align}
We randomly draw task pairs and annotate win-lose-tie for each metric. Then we compute the average score of each task and rank all the tasks. 

Win-lose-tie annotation is more time-consuming than a simple scoring for each task. We solve this problem by introducing a hybrid task filtering and task comparison mechanism. If we want to get an expectation of $k$ comparison scores for each of the $n$ tasks, we have to make $kn/2$ comparisons. We add a feature that can mark a task meaningless, meaning we believe that it cannot be more meaningful than any other task. Assume we mark $m$ meaningless tasks, the number of comparisons can be reduced to $k(n-m)/2$. Usually, we can remove more than half of the prediction tasks in this way. As soon as a task is marked meaningless, it will not be used for comparison. Table \ref{tab:ann} shows the number of annotations we got.

\begin{figure*}[t]
     \centering
     \includegraphics[width=.96\textwidth]{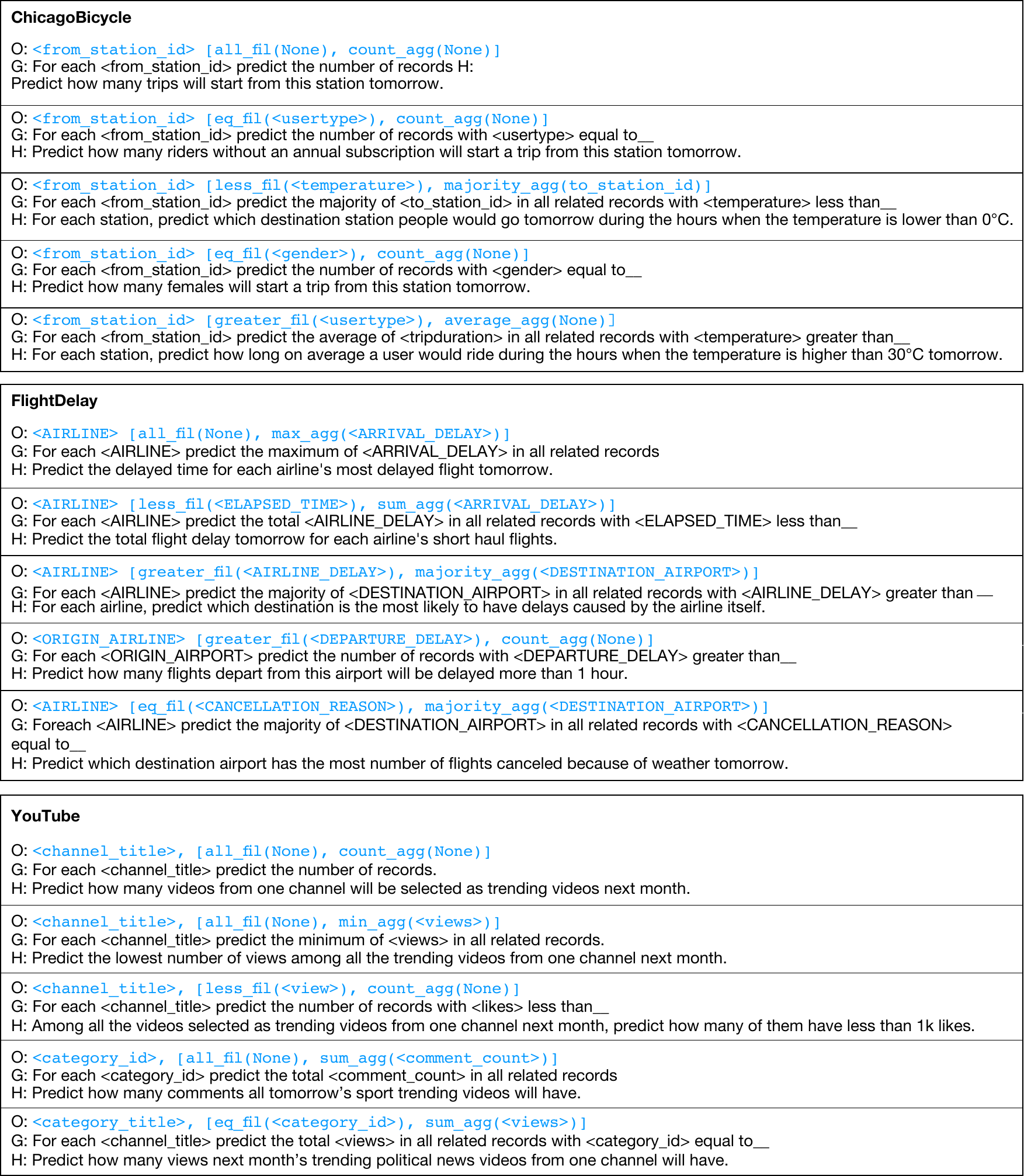}
     \caption{Top rated prediction tasks on Chicago Bicycle, Flight Delay, and YouTube Trending dataset. O is the structural representation of the task, including entity and a sequence of operations. G is the generated description of the task. H is the human interpretation of the task. }
     \label{fig:case}
\end{figure*}

\begin{figure*}[t]
    \centering
    \begin{subfigure}[t]{0.3\textwidth}
    \centering
    Chicago Bicycle\\
    \includegraphics[width=\textwidth]{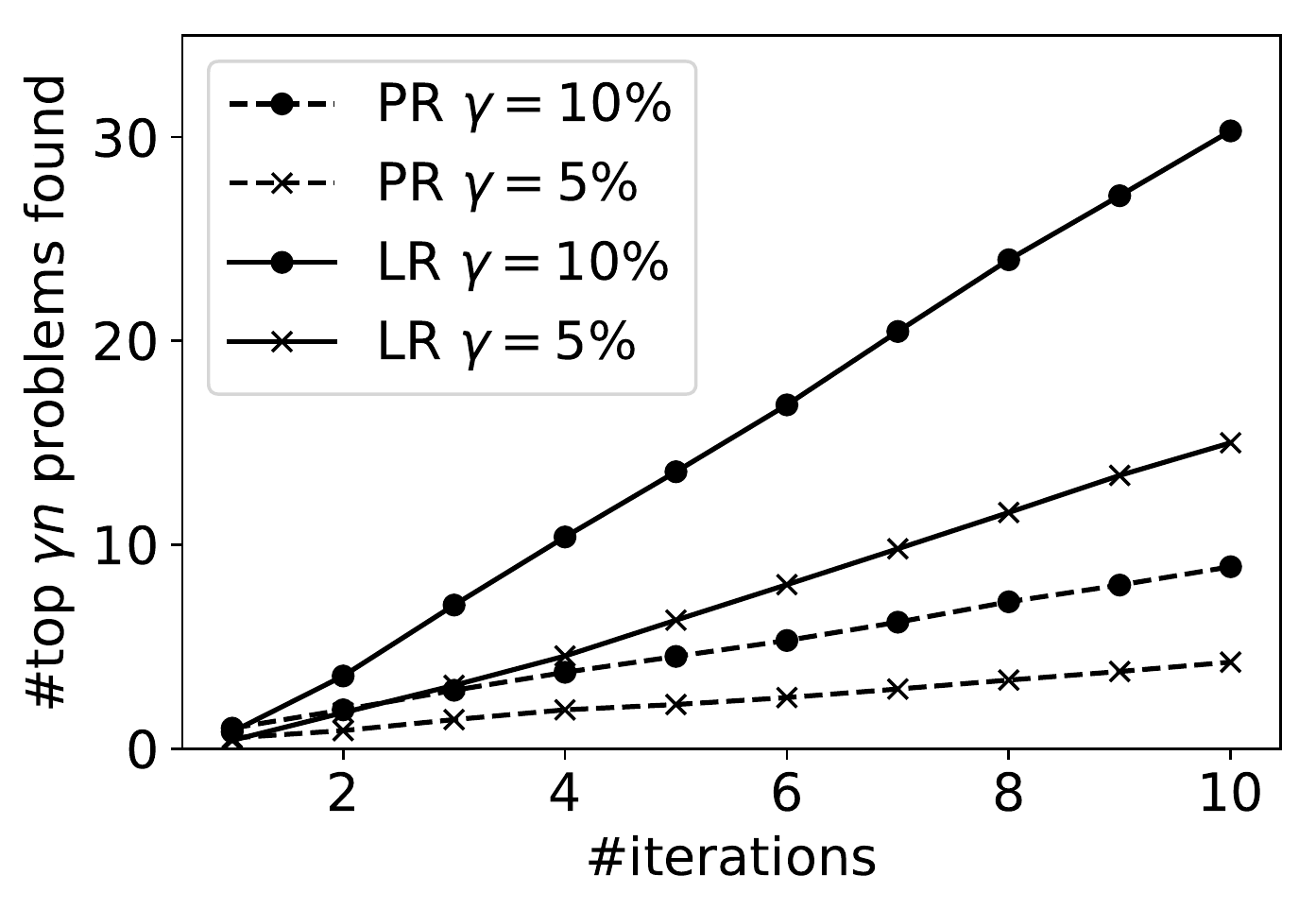}
    \end{subfigure}
    \begin{subfigure}[t]{0.3\textwidth}
    \centering
    Flight Delay\\
    \includegraphics[width=\textwidth]{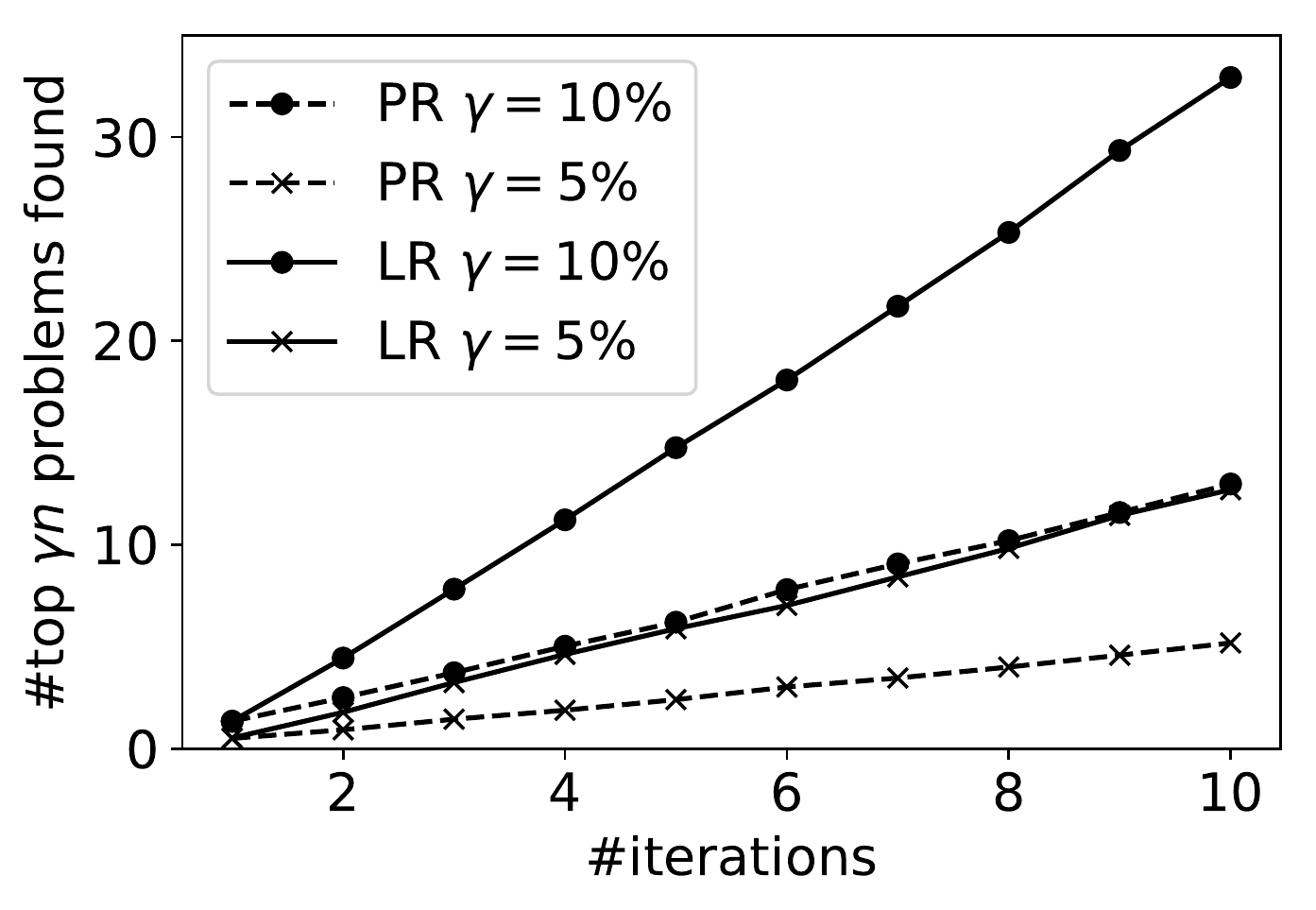}
    \end{subfigure}
    \begin{subfigure}[t]{0.3\textwidth}
    \centering
    YouTube Trending\\
    \includegraphics[width=\textwidth]{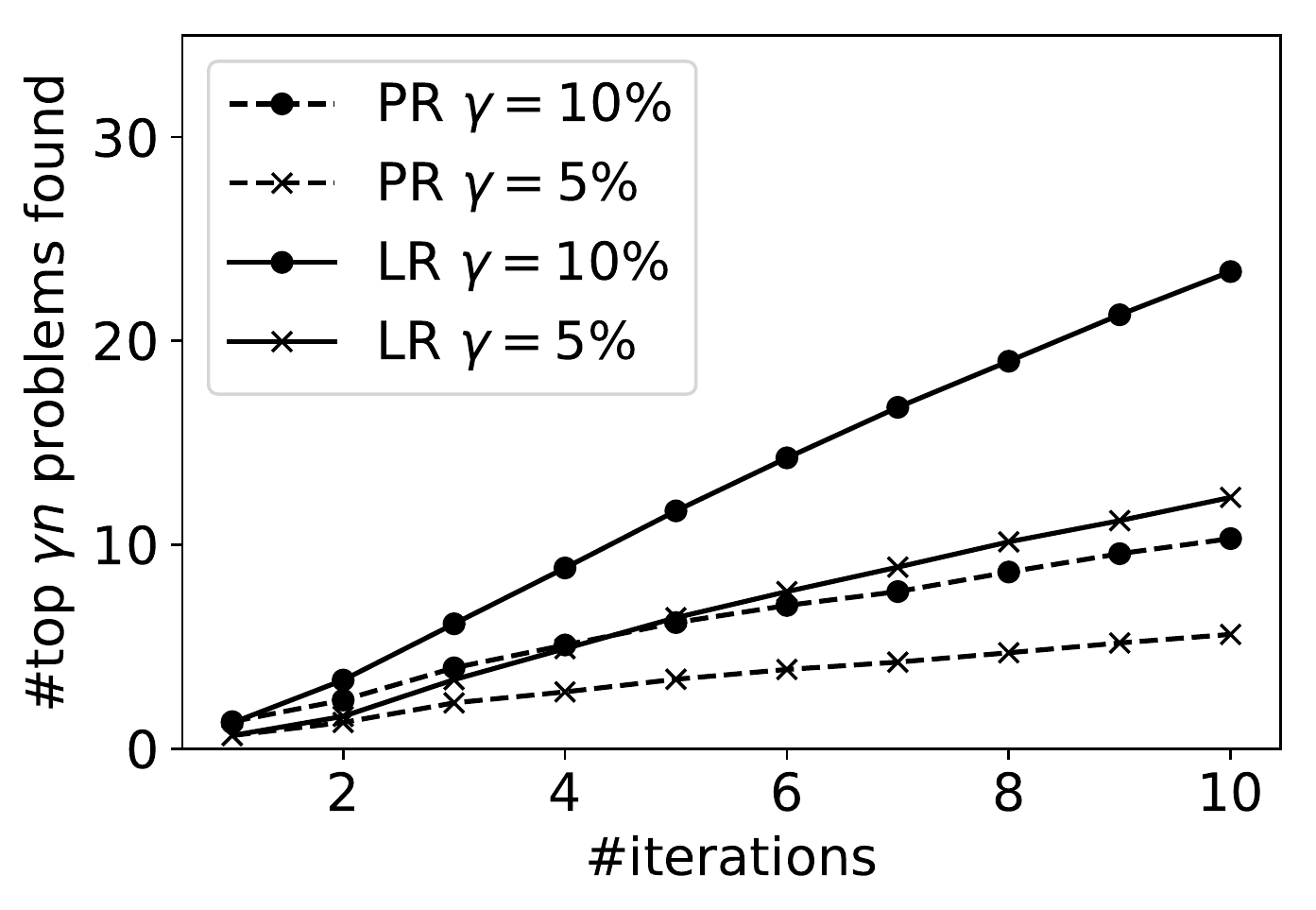}
    \end{subfigure}
    \caption{Interactive recommendation simulation results. We compare our model (LR) and uniform selection (PR) on three datasets. For each method, we evaluate how many top-10\% tasks ($\gamma=10\%$) and top-5\% tasks ($\gamma=5\%$) are found in 10 interactive iterations. We run experiment $100$ times and plot the average. We conducted the \texttt{t-test} for each of the comparisons between \texttt{PR} and \texttt{LR} and found the significance level to be lower than or equal to $0.05$.}
    \label{fig:rec}
\end{figure*}

\section{Discussion}\label{sec:disc}

\noindent \textbf{Did our prediction task generation algorithm generate meaningful problems?}: Our generation procedure created $99$, $415$, and $208$ meaningful prediction tasks on Chicago Bicycle, Flight Delay, and YouTube Trending respectively, showing that our task space has good coverage. This demonstrates that it can find many useful prediction tasks on a variety of datasets. Roughly $23.6\%$, $20.0\%$, and $53.1\%$ of the tasks were meaningful for these datasets. Considering the fact that formulating a useful prediction task is challenging for experts, our \texttt{MLFriend}, with good coverage and usefulness, is helpful for data scientists. 

\noindent \textbf{What are the reasons for creating meaningless problems?}: A number of meaningless problems are created since the system simply does not take into account the semantic meaning of columns. For example, the Chicago Bicycle dataset has a low meaningful rate, since the system tries to create prediction tasks that are about predicting the number of docks in origin and destination stations (which are not likely to change). A similar issue happens on Flight Delay dataset. Meaningless tasks generated on the flight delay dataset tried to predict the number of scheduled flights. 

We also find some tasks annotated as meaningless are in fact useful when read from a different perspective. For example, \textit{predict the number of long-haul fights next year} is meaningless if the user is interested in delay and cancellation, but it is actually useful if the user tries to study transportation growth. 

This analysis opens door to a fascinating area of research that would create task filters based on natural language understanding. 

\noindent \textbf{How accurate are the machine learning models for these tasks?}: Fig. \ref{fig:hist} shows the histogram of $R^2$ for all regression tasks and the histogram of accuracy for all classification tasks. For regression tasks, lots of tasks get $0.9 < R^2 < 1.0$. Such high $R^2$ indicates trivial prediction tasks. For tasks with $R^2<0.9$, the distribution varies between different datasets. For classification tasks, there's no clear pattern for accuracy distribution, because making predictions on different datasets involves different levels of difficulty.

On Fig. \ref{fig:hist}, we also show the distribution of meaningful prediction tasks over performance. We observe that $21.4\%$ percent of regression tasks and $41.5\%$ percent of classification tasks are meaningful. In regression tasks, those with $R^2 > 0.9$ tend to be trivial and meaningless, while classification tasks with high accuracy do not have the same phenomenon. Using our predefined operation sets, classification tasks are more meaningful then regression tasks. 

\noindent \textbf{What are the top rated tasks and are they useful?}: Fig. \ref{fig:case} shows top rated tasks in each dataset. We show structured representation, generated natural language and human interpretation in the table. Despite using only two successful operations to generate labels, these top-rated prediction tasks are very useful when interpreted by different users. 
\begin{itemize}
    \item On the Chicago Bicycle dataset, prediction tasks generated for each origin station are more interesting than prediction tasks generated for all the data. For example predicting majority destination and/or the number of trips made from each station can help the bike sharing company optimize their operation, while predicting the number of daily pass users in each station can help determine the type of ads that should be placed there. 
    \item On Flight Delay dataset, \texttt{MLFriend} can formulate tasks helping airline and airport operation. For example, airports can predict how many flights will be delayed so that they can get prepared.
    \item On the YouTube dataset, \texttt{MLFriend} generates interesting predictions for both channels and categories. The prediction for the number of views and comments can help channels operate efficiently and properly prepare computing resources.
\end{itemize}

\noindent \textbf{Did automatic problem discovery approach work?}: To test the recommendation system, we use our annotations to simulate a real user. In each step, the recommendation system is allowed to show 10 prediction tasks to a user, whereupon the user gives feedback. The recommendation system interacts with data scientist for 10 iterations. We evaluate how many top-10\% ($\gamma=10\%$) and top-5\% ($\gamma=5\%$) prediction tasks are shown to the user during this process. 

We simulate the user using the following model. We assume a user is more likely to mark a prediction task "good" if a task has a higher rank in our annotation, and that he will not mark a really bad task as "good." Specifically, they will mark a task "good" with a probability 
\begin{equation}
    \mathbb{P} = 
    \begin{cases}
        1 - \frac{r_1}{N}, & \text{ if $r_1$ < $\frac{N}{2}$},\\
        0, & \text{ otherwise}.
    \end{cases}
\end{equation}
where $r_i$ is the rank of the prediction task in our annotation. Since there is no existing baseline, we compare our model with uniform selection (PR). We run this experiment 100 times to avoid randomness. 

 Fig. \ref{fig:rec} plots the average number of top prediction tasks found. We observe that our interactive recommendation algorithm significantly and consistently outperforms uniform selection in each iteration and on all three datasets. With this we show that it is possible to recommend prediction tasks. With a simple linear model, our algorithm can discover 2 to 3 times more top-rated tasks comparing with uniform selection.

%% file: datasetdetail.tex
\begin{table}[htb]
\begin{tabular*}{\linewidth}{@{\extracolsep{\fill}}lcccc}\toprule
Dataset          & \#Entity & \#Categorical & \#Numerical  \\\midrule
Chicago Bicycle  & 2        & 3             & 4          \\
Flight Delay     & 4        & 4             & 10      \\
YouTube Trending & 1        & 1             & 4      \\\bottomrule
\end{tabular*}
\caption{Number of entity columns, categorical columns and numerical columns in each dataset. Total number of columns is \#Entity+\#Categorical+\#Numerical+1(time column). }
\label{tab:ncol}
\end{table}

%% file: 6.conclusion.tex
\section{Conclusion and Future Work}\label{concludes}
In this paper, we tackle a fundamental challenge in automated data science: automatically discovering interesting, meaningful prediction problems on event-driven time series data. We define a clear prediction problem representation, then build \texttt{MLFriend}, a system that generates prediction tasks, learns and ranks them using a \textit{meta model} and solves prediction tasks using \texttt{AutoML} frameworks. By applying our system to 3 datasets, we show that our system can properly define and solve prediction tasks. We further show that it's possible to build a user-specific recommendation system to suggest useful problems. 

In the future, we will pursue the following directions. 
\begin{itemize}
    \item The task space is still relatively small -- for example, it does not support multiple column operations. We will study how to introduce these features without increasing the task space too much.
    \item The natural language description is still not understandable for non-experts. We will try to annotate data and build a neural model to generate high-quality descriptions for non-experts.
    \item We want to figure out how to effectively annotate large-scale prediction problems to train a better recommendation system. 
\end{itemize}